\documentclass[conference,10pt,a4paper]{IEEEtran}
\IEEEoverridecommandlockouts
\usepackage{cite}
\usepackage{amsmath,amssymb,amsfonts}
\usepackage{algorithmic}
\usepackage{graphicx}
\usepackage{textcomp}
\usepackage{xcolor}
\def\BibTeX{{\rm B\kern-.05em{\sc i\kern-.025em b}\kern-.08em
    T\kern-.1667em\lower.7ex\hbox{E}\kern-.125emX}}
\begin{document}

\title{AttriGen: Automated Multi-Attribute Annotation for
Blood Cell Datasets}

\author{
  \IEEEauthorblockN{%
    Walid Houmaidi\IEEEauthorrefmark{1}, 
    Youssef Sabiri\IEEEauthorrefmark{1}, 
    Fatima Zahra Iguenfer, 
    Amine Abouaomar
  }
  \IEEEauthorblockA{%
    School of Science and Engineering, Al Akhawayn University\\
    Ifrane, Morocco\\
    Email: \{w.houmaidi, y.sabiri, f.iguenfer, a.abouaomar\}@aui.ma
  }%
  \thanks{* Walid Houmaidi and Youssef Sabiri contributed equally to this work and are co-first authors.}
}

\maketitle

\begin{abstract}
We introduce AttriGen, a novel framework for automated, fine-grained multi-attribute annotation in computer vision, with a particular focus on cell microscopy where multi-attribute classification remains underrepresented compared to traditional cell type categorization. Using two complementary datasets: the Peripheral Blood Cell (PBC) dataset containing eight distinct cell types and the WBC Attribute Dataset (WBCAtt) that contains their corresponding 11 morphological attributes, we propose a dual-model architecture that combines a CNN for cell type classification, as well as 
a Vision Transformer (ViT) for multi-attribute classification achieving a new benchmark of 94.62\% accuracy. Our experiments demonstrate that AttriGen significantly enhances model interpretability and offers substantial time and cost efficiency relative to conventional full-scale human annotation. Thus, our framework establishes a new paradigm that can be extended to other computer vision classification tasks by effectively automating the expansion of multi-attribute labels.
\end{abstract}


\section{Introduction}
\label{sec:intro}

Leukemia caused 309{,}006 deaths in 2018, ranking 11\textsuperscript{th} among cancer mortalities worldwide \cite{statistic}. Early diagnosis hinges on microscopic review of blood smears, a task that is slow, labor-intensive, and increasingly hampered by shortages of laboratory experts \cite{diagnosis,MedicalShortage}. Although digital pathology promises relief, most automated systems still focus on coarse cell‐type classification, overlooking the fine-grained morphological cues that guide clinical decisions \cite{CNNTypeClass}.

We bridge this gap with a \textbf{dual-model architecture} that mirrors the pathologist’s workflow:  
(1) a CNN trained on the PBC dataset \cite{PBC} distinguishes eight leukocyte classes, and  
(2) a Vision Transformer (ViT) trained on WBCAtt \cite{WBCAtt} predicts eleven expert-defined attributes.  
The fused output yields a 12-attribute profile—cell type plus eleven morphology traits—providing a richer, clinic-aligned characterization.

To scale this capability, we introduce \textbf{AttriGen}, an automated annotation pipeline that applies the dual model to unlabeled images, producing high-quality multi-attribute labels with minimal human intervention. At 94.62\% accuracy—near human expert performance—AttriGen annotates datasets orders of magnitude faster than manual review, allowing rapid dataset expansion and accelerating downstream model development.

Our main contributions are:
\begin{itemize}
    \item A hybrid CNN–ViT framework that unifies cell-type and attribute recognition for comprehensive WBC analysis.
    \item AttriGen, an automated pipeline that generates clinically consistent, multi-attribute annotations at scale.
    \item An empirical study of accuracy–time trade-offs, demonstrating substantial efficiency gains without clinically significant loss in precision.
\end{itemize}

\section{Related Work}
\label{sec:formatting}

\subsection{Blood Cell Analysis}
Microscopic review of peripheral blood smears is essential in hematology, yet manual inspection remains slow, costly, and dependent on scarce experts \cite{ClinicalMethods,diagnosis}. Because WBC morphology signals infections and malignancies, reliable automated analysis is a long‐standing goal \cite{hoffbrand2019essential}.

\subsection{Evolution of Automated Blood Cell Analysis}

\textbf{Early pipelines.} Initial systems chained preprocessing, segmentation, hand-crafted feature extraction, and shallow classifiers \cite{sabino2004texture,WBCSegmentation}. Although foundational, they struggled with staining and imaging variability.

\noindent\textbf{CNN era.} Deep learning removed the need for manual features: transfer-learned CNNs achieved 96\% accuracy on eight blood cell types \cite{CNNTypeClass}, while lightweight BloodCell-Net reached 97\% across nine types after U-Net segmentation \cite{BloodCellNet}.

\subsection{Attribute Recognition in Medical Imaging}
Attribute-based recognition decomposes images into interpretable traits \cite{lampert2009learning}. Beyond hematology, it has improved lung CT classification \cite{cheplygina2019notso}. In blood cells, progress was limited until the WBCAtt dataset with 11 morphological attributes became available \cite{WBCAtt}. WBCAtt enabled P4BC (ACAcc 63.6 \%) and ConceptCLIP (AUC 99.5\%) \cite{PartialLabelLearning,ConceptCLIP}. Earlier datasets such as PBC offered only coarse cell labels \cite{PBC}.

\subsection{Multi-Model Medical Imaging}
Combining models boosts robustness. Ensembles of pre-trained networks outperform single models in breast histopathology \cite{litjens2017survey}, while multi-task CNNs simultaneously detect regions and classify bone‐marrow cells, achieving 97\% ROI accuracy \cite{Histogram}.

\begin{figure*}[!t]
    \centering
    \includegraphics[width=\textwidth]{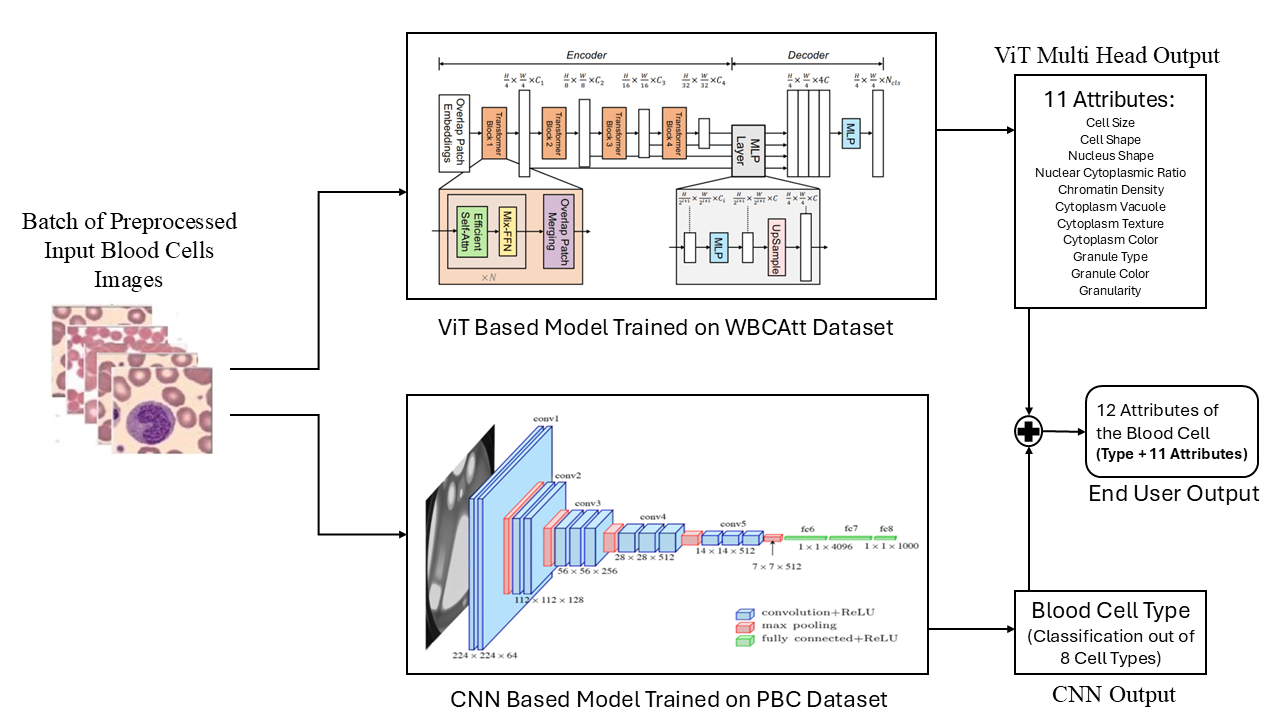}
    \caption{System Architecture}
    \label{fig:system_architecture}
\end{figure*}

\subsection{Automated Dataset Annotation}
Modern annotation pipelines cut labeling cost through semi-supervision and synthetic data. FixMatch attains 94.9\% on CIFAR-10 with 250 labels by pseudo-labeling \cite{FixMatch}. Domain-randomized synthetic images further reduce manual effort and rival real-data training \cite{TainingDeepNet}. 
Despite these advances, such techniques have been mostly applied to single-label classification tasks, with limited exploration in multi-attribute classification. This presents a significant challenge for domains like cell microscopy, where comprehensive morphological characterization requires multiple interdependent attributes to be annotated simultaneously; a gap that our proposed AttriGen framework addresses.

\section{Methodology}
\label{sec:methodology}

\subsection{System Architecture}

Our system employs a dual-model architecture that processes blood cell images through two parallel pathways, as illustrated in Figure \ref{fig:system_architecture}. This design leverages the complementary strengths of different deep learning architectures while capitalizing on the specialized information available in two distinct datasets. Given a batch of preprocessed blood cell images, our system performs two parallel inference operations:

\noindent{\textbf{Morphological Attribute Recognition}:} A ViT-based model, trained on the WBCAtt dataset \cite{WBCAtt}, analyzes the input images to extract 11 fine-grained morphological attributes. These attributes include cell size, cell shape, nucleus shape, nuclear-cytoplasmic ratio, chromatin density, cytoplasm texture, cytoplasm color, cytoplasm vacuole, granularity, granule type, and granule color. The ViT architecture is particularly effective for this task due to its ability to model long-range dependencies and complex relationships between different cellular components through its self-attention mechanisms.

\noindent{\textbf{Cell Type Classification}:} Concurrently, a CNN model, trained on the larger PBC dataset \cite{PBC}, classifies the input images into one of eight cell types: neutrophils, eosinophils, basophils, lymphocytes, monocytes, immature granulocytes, erythroblasts, and platelets. The CNN architecture excels at capturing hierarchical features and local patterns critical for cell type differentiation.

The outputs from both models are combined to create a comprehensive 12-attribute characterization of each blood cell image. This integration mimics the holistic assessment process used by clinical pathologists, who consider both cell type and morphological characteristics when making diagnostic decisions. 

\subsection{Datasets Used}
\subsubsection{Description of the Datasets}

Our study relies on two complementary WBC image sets.

\textbf{PBC} \cite{PBC} — 17,092 May Grünwald-Giemsa–stained single-cell images (360 × 363 px, JPG) captured by a CellaVision DM96 from healthy donors. They cover eight classes: neutrophils (3,329), eosinophils (3,117), basophils (1,218), lymphocytes (1,214), monocytes (1,420), immature granulocytes

\begin{figure*}[ht]
    \centering
    \includegraphics[width=\textwidth]{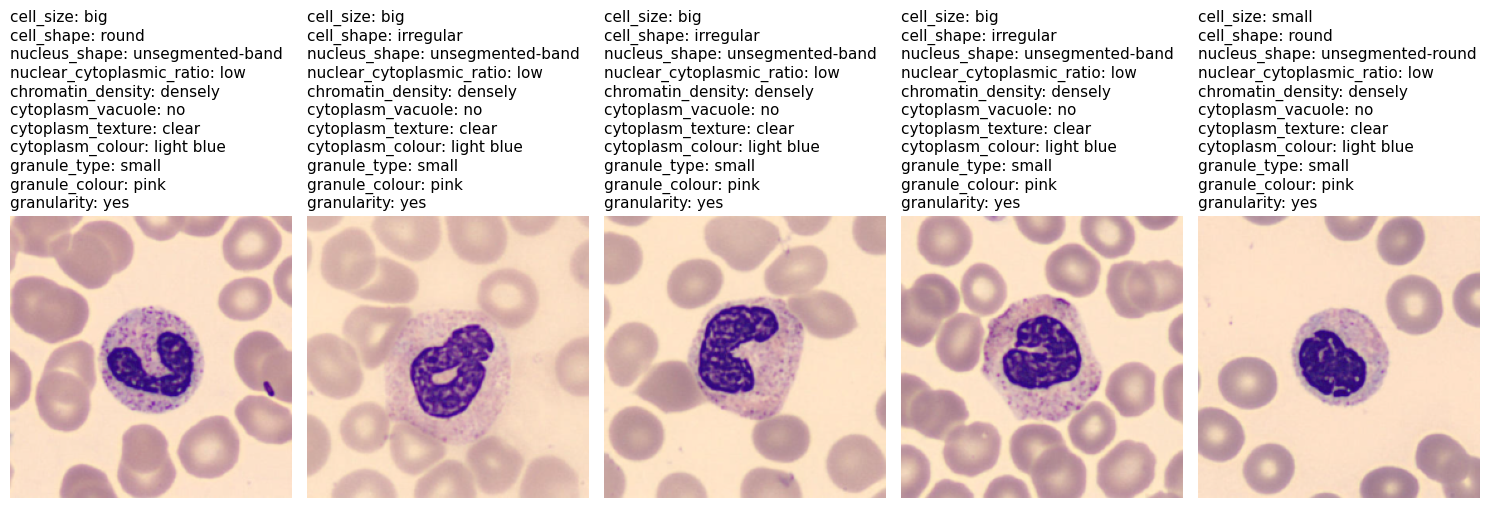}
    \caption{Sample images from the WBCAtt dataset}
    \label{fig:wbcatt_sample}
\end{figure*}

(2,895), erythroblasts (1,551) and platelets (2,348). We use them to train the CNN cell-type classifier.

\textbf{WBCAtt} \cite{WBCAtt} — 10,298 images annotated with 11 expert-defined morphological attributes, spanning cell and nucleus size/shape, N:C ratio, chromatin density, cytoplasm texture / colour / vacuoles, and granule presence, type and colour (see Fig.~\ref{fig:wbcatt_sample}). These labels drive the ViT model for fine-grained attribute recognition.

\subsubsection{Preprocessing Steps}

\noindent{\textbf{PBC Dataset Preprocessing:}} We developed a robust preprocessing pipeline for the PBC dataset. First, we standardized images by resizing them to 224$\times$224 pixels, converting them to NumPy arrays, and normalizing pixel values between [0--1] for stable gradient computations during training. Furthermore, we converted textual labels into numerical values and applied one-hot encoding, making them suitable for CNN-based multi-class classification while maintaining categorical relationships.
\newline
\noindent{\textbf{WBCAtt Dataset Preprocessing:} We performed advanced preprocessing for the multi-attribute WBCAtt dataset. Initially, CSV files with image paths and attribute annotations were loaded. Attributes were standardized through alphabetically sorted bijective mappings between textual labels and numeric encodings. Images were resized to 224$\times$224 pixels, normalized, and converted to tensors. Data were serialized into dictionaries with preprocessed images and encoded labels. This streamlined pipeline enhanced model performance and training efficiency.

\subsubsection{Data Splits}
We applied standardized dataset splits for robust evaluation. The PBC dataset was divided into 80\% training (13,673 images), 10\% validation (1,710 images), and 10\% test data (1,709 images), ensuring balanced representation. For the WBCAtt dataset, splits comprised approximately 60\% training (6,179 images), 10\% validation (1,030 images), and 30\% test data (3,099 images), effectively capturing attribute variations and enabling thorough performance assessment.

\subsubsection{Data Augmentation} 
\noindent{\textbf{PBC Dataset Augmentation:}
\begin{itemize}
    \item Shear Transformation set to 0.3 to introduce shape distortions.
    \item Zoom set to 0.3 for simulating scale variations.
\end{itemize}

\noindent{\textbf{WBCAtt Dataset Augmentation for VIT based models:}
\begin{itemize}
\item Random Horizontal Flip to enhance orientation diversity.
\item Resize (256 pixels, maintaining aspect ratio) along with Center Crop (224×224 pixels) for standardized evaluation.
\end{itemize}

\noindent{\textbf{WBCAtt Dataset Augmentation for CNN Model:}
\begin{itemize}
\item Random Crop (224×224 pixels) to improve generalization by focusing on different image regions.
\item Random Horizontal Flip to enhance robustness through orientation variations. 
\end{itemize}
\subsection{Models Used}
Our AttriGen framework leverages specialized deep learning architectures for the distinct tasks of cell type classification and morphological attribute recognition.
\newline
\noindent{\textbf{Cell Type Classification:} For classifying blood cells into eight types, we implemented the VGG16 architecture \cite{VGG} pre-trained on ImageNet and fine-tuned on the PBC dataset. VGG's hierarchical convolutional structure effectively captures the local features and spatial patterns critical for distinguishing between different cell types.
\newline
\noindent{\textbf{Architectures for Attribute Recognition:} We evaluated multiple CNN and ViT architectures, selecting the following best performing models:
\begin{itemize}
\item \textbf{VGG16} \cite{VGG}: A CNN known for strong feature extraction and excellent performance in visual recognition tasks.
\item \textbf{ViT-B/16} \cite{dosovitskiy2021image}: Utilizes 16×16 image patches processed through transformer encoders to effectively capture global cell relationships.
\item \textbf{Swin-S} \cite{SwinB}: Features hierarchical shifted-window structures enabling efficient multi-scale feature extraction.
\item \textbf{DeiT-B} \cite{touvron2021training}: Employs knowledge distillation for efficient yet high-performance training.
\item \textbf{DeiT-S} \cite{touvron2021training}: Compact, parameter-efficient variant suitable for resource-constrained scenarios.
\end{itemize}

\subsection{Results}
\subsubsection{PBC Results}

\begin{table}[h]
    \centering
    \renewcommand{\arraystretch}{1}
    \setlength{\tabcolsep}{4pt}
    \caption{Comparison of different models and their accuracy on PBC dataset}
    \resizebox{\columnwidth}{!}{%
    \begin{tabular}{|l|l|c|}
        \hline
        \textbf{Papers} & \textbf{Methods Used} & \textbf{Accuracy (\%)} \\
        \hline
        \cite{CNN99.91} & Custom CNN-based Framework & 99.91 \\
        \cite{CNN99.72} & Multibranch Lightweight CNN & 99.72 \\
        \cite{EfficientNet99.69} & Attention-based Data Augmentation (EfficientNet) & 99.69 \\
        \cite{Swim+ConvMixer} & SC-MP-Mixer (Swin Transformer + ConvMixer) & 99.65 \\
        \cite{RCNN+CNN} & R-CNN + CNN Classification & 99.31 \\
        \cite{BloodCaps} & BloodCaps (Capsule Network) & 99.30 \\
        \cite{modifiedCNN} & Modified Inception-based CNN & 98.89 \\
        Our Model & VGG16 & \textbf{98.83} \\
        \cite{BCNet} & BCNet (Transfer Learning + Ensemble) & 98.51 \\
        \cite{ShuffleNet} & ShuffleNet-based CNN & 97.94 \\
        \hline
    \end{tabular}
    }
    \label{tab:pbc_accuracy}
\end{table}

The Table \ref{tab:pbc_accuracy} shows comparative results on PBC dataset. Although our model did not achieve the global benchmark, as obtaining the highest possible accuracy was not the primary focus of our study, it still demonstrated strong performance, reaching an accuracy of 98.83\%. This result highlights the effectiveness and potential of our proposed methodology within the scope of the research.

\subsubsection{Metrics Used}
To evaluate the performance of our classification model across multiple attributes, we utilized five key evaluation metrics:
\newline
\noindent{\textbf{Predefined Metrics:}
\begin{itemize}
    \item \textbf{Accuracy (\(Acc\))}: Measures the proportion of correctly classified samples out of the total samples.
    \item \textbf{Precision (\(P\))}: Represents the proportion of correctly predicted positive samples out of all predicted positives.
    \item \textbf{Recall (\(R\))}: Also known as sensitivity or true positive rate, it quantifies the proportion of actual positive samples that were correctly predicted.
    \item \textbf{F1-score (\(F1\))}: A harmonic mean of precision and recall, providing a balance between the two metrics.
\end{itemize}
\noindent{\textbf{Global Average Accuracy (GAA):} To provide a holistic evaluation across multiple attributes, we define a \textbf{GAA} metric. This is computed by averaging the accuracy values across all attribute heads:
 
\begin{equation}
Acc_{global} = \frac{1}{N} \sum_{i=1}^{N} Acc_i
\end{equation}

where \(N\) is the total number of attribute heads, and \(Acc_i\) is the accuracy of the \(i^{th}\) attribute head. This metric ensures that the overall performance of the model across different attributes is effectively represented.
\newline
Unlike previous studies that employed metrics such as AUC \cite{ConceptCLIP} or ACAcc\cite{PartialLabelLearning}, we chose GAA due to its straightforward interpretability and its capacity to equally weigh each attribute prediction without bias toward any specific attribute class.

\subsubsection{WBCAtt Results}
\begin{table}[h]
    \centering
    \caption{Comparison of different models and their GAA on multi-attribute prediction on WBCAtt}
    \resizebox{\columnwidth}{!}{%
    \begin{tabular}{|l|l|c|}
        \hline
        \textbf{Authors} & \textbf{Methods Used} & \textbf{GAA(\%)} \\
        \hline
        Our Model & Swin-S & \textbf{94.62} \\
        Our Model & ViT-B/16 & \textbf{94.35} \\
        Our Model & DeiT-B & \textbf{94.31} \\
        Our Model & DeiT-S & \textbf{94.19} \\
        Our Model & VGG16 & \textbf{91.42} \\
        \cite{DAFNET} & Morphological Attributes Predictor (MAP) & 88.65 \\
    
        \hline
    \end{tabular}
    }
    \label{tab:attribute_accuracy}
\end{table}
The Table~\ref{tab:attribute_accuracy} shows comparative results on WBCAtt dataset. Our Swin-S model established a new global benchmark by achieving a GAA of 94.62\%.

\subsubsection{Results Comparison}
\begin{table}[h]
    \centering
    \renewcommand{\arraystretch}{1.2}
    \setlength{\tabcolsep}{4pt}
    \caption{Classification performance metrics breakdown for each attribute of Swin-S Model}
    \resizebox{\columnwidth}{!}{%
    \begin{tabular}{|l|c|c|c|c|}
        \hline
        \textbf{Attribute} & \textbf{Accuracy (\%)} & \textbf{Precision (\%)} & \textbf{Recall (\%)} & \textbf{F1 Score (\%)} \\
        \hline
        Cell Size & 84.03 & 84.02 & 84.03 & 83.99 \\
        Cell Shape & 94.29 & 94.40 & 94.29 & 94.33 \\
        Nucleus Shape & 80.51 & 80.69 & 80.51 & 80.34 \\
        Nuclear-Cytoplasmic Ratio & 98.61 & 98.60 & 98.61 & 98.60 \\
        Chromatin Density & 94.93 & 94.71 & 94.93 & 94.79 \\
        Cytoplasm Vacuole & 97.58 & 97.59 & 97.58 & 97.59 \\
        Cytoplasm Texture & 96.55 & 96.65 & 96.55 & 96.58 \\
        Cytoplasm Colour & 95.58 & 95.60 & 95.58 & 95.57 \\
        Granule Type & 99.61 & 99.61 & 99.61 & 99.61 \\
        Granule Colour & 99.29 & 99.29 & 99.29 & 99.29 \\
        Granularity & 99.81 & 99.81 & 99.81 & 99.81 \\
        \hline
        \textbf{Global Average} & \textbf{94.62} & \textbf{94.63} & \textbf{94.62} & \textbf{94.59} \\
        \hline
    \end{tabular}
    }
    \label{table:metrics}
\end{table}

\begin{figure}[ht]
    \centering
    \includegraphics[width=\columnwidth]{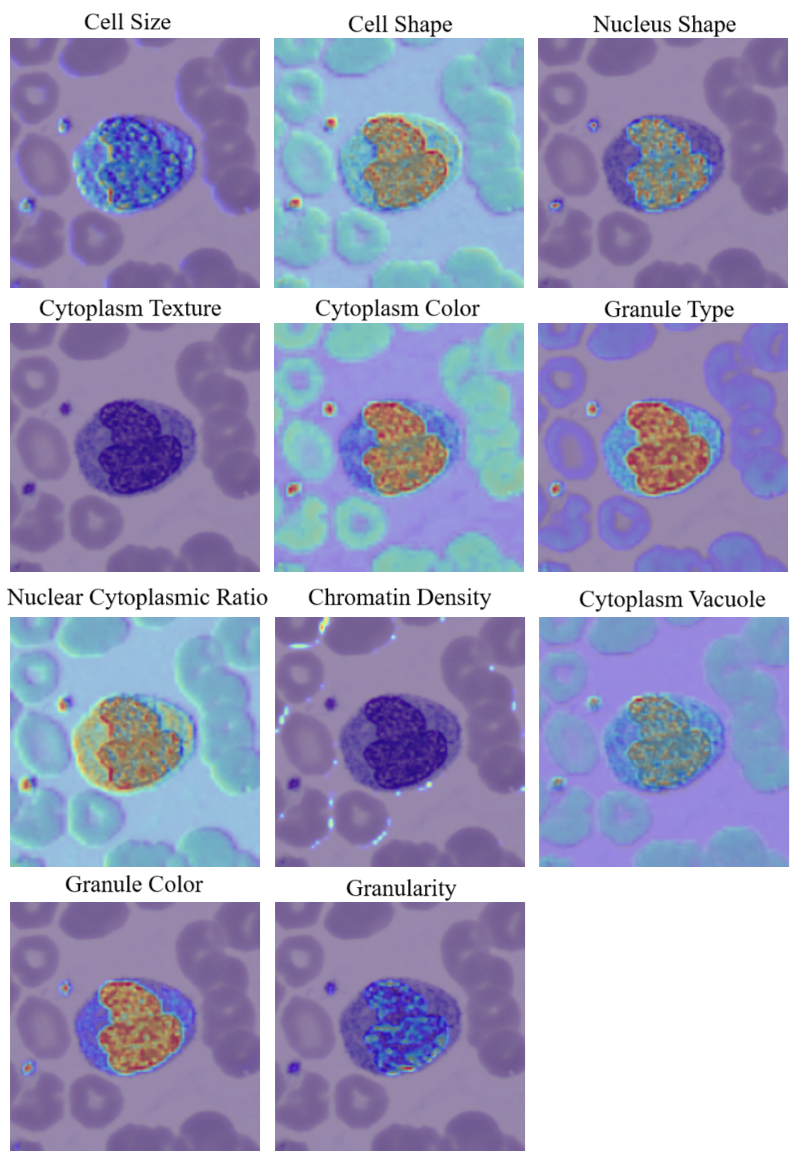}
    \caption{Grad-CAM For Swin-S Model Correct Predictions on a Random Image from Test Set}
    \label{fig:gradcam}
\end{figure}

Among the tested models on WBCAtt Dataset, Swin-S achieved the highest accuracy at \textbf{94.62\%} (Table~\ref{tab:attribute_accuracy}), closely followed by other ViT variants, confirming the effectiveness of transformer-based models in attribute recognition tasks. VGG16, with an accuracy of 91.42\%, demonstrated impressive performance considering its lower computational demand, making it suitable for efficiency-focused scenarios. All five models surpassed previous benchmarks, underscoring their robust performance across different computational requirements.

Table~\ref{table:metrics} provides a detailed breakdown of Swin-S’s performance across different attributes. The model performed exceptionally well in recognizing \textbf{Granularity, Granule Type, and Granule Colour}, achieving near-perfect accuracy \textbf{(99.81\%, 99.61\%, and 99.29\%, respectively)}. These results suggest that transformer-based models can effectively capture fine-grained cellular features.

Attributes like \textbf{Nuclear-Cytoplasmic Ratio (98.61\%) and Cytoplasm Vacuole (97.58\%)} also showed high classification performance, indicating that the model is adept at distinguishing nuclear and cytoplasmic characteristics. However, Nucleus Shape (80.51\%) and Cell Size (84.03\%) had relatively lower accuracy, suggesting room for improvement in learning shape-related morphological variations.

To better understand the decision-making process of Swin-S, we employed \textbf{Grad-CAM} visualization, as shown in Figure~\ref{fig:gradcam}. This highlights the most important regions in the image that contributed to the model's predictions, allowing for better interpretability of the attribute classification results.

\subsection{AttriGen}

Our dual-model architecture demonstrates strong performance in both cell type classification and morphological attribute recognition tasks. However, the broader application of such attribute-based approaches in medical image analysis remains constrained by a critical limitation: the scarcity of comprehensively annotated datasets. The WBCAtt dataset exemplifies this challenge, as its creation involved a meticulous and labor-intensive annotation process. Specifically, biomedical students initially annotated over 10,000 WBC images, each with 11 morphological attributes. Subsequently, research scientists meticulously reviewed each image and its assigned attributes, ensuring that every image was inspected by at least two individuals. Ambiguities were resolved through discussions with pathologists. This complex process is both time-consuming and costly, as it requires specialized domain knowledge, making the creation of large-scale multi-attribute datasets prohibitively expensive \cite{WBCAtt}. This data bottleneck represents a significant obstacle to advancing interpretable AI systems in hematological diagnosis and similar medical domains. It is imperative to explore more efficient and scalable annotation methods to overcome these challenges?

\subsubsection{Proposed Approach} 
To address the challenges associated with manual annotation of WBC images, we introduce AttriGen, an automated system designed to efficiently and accurately annotate multi-attribute datasets. Considering the PBC dataset, which comprises 17,092 images, manually annotating each image would be a formidable task. In contrast, applying our solution to the remaining 6,784 unannotated cells in the PBC dataset at a rate of 20 milliseconds per cell would result in approximately 2.26 minutes, a considerable improvement in efficiency. This automated approach not only accelerates the annotation process but also minimizes the reliance on specialized domain knowledge, thereby reducing costs and facilitating the development of large-scale, comprehensively annotated datasets.

\subsubsection{AttriGen: Framework Architecture} 
\begin{figure}[h]
    \centering
    \includegraphics[width=\columnwidth]{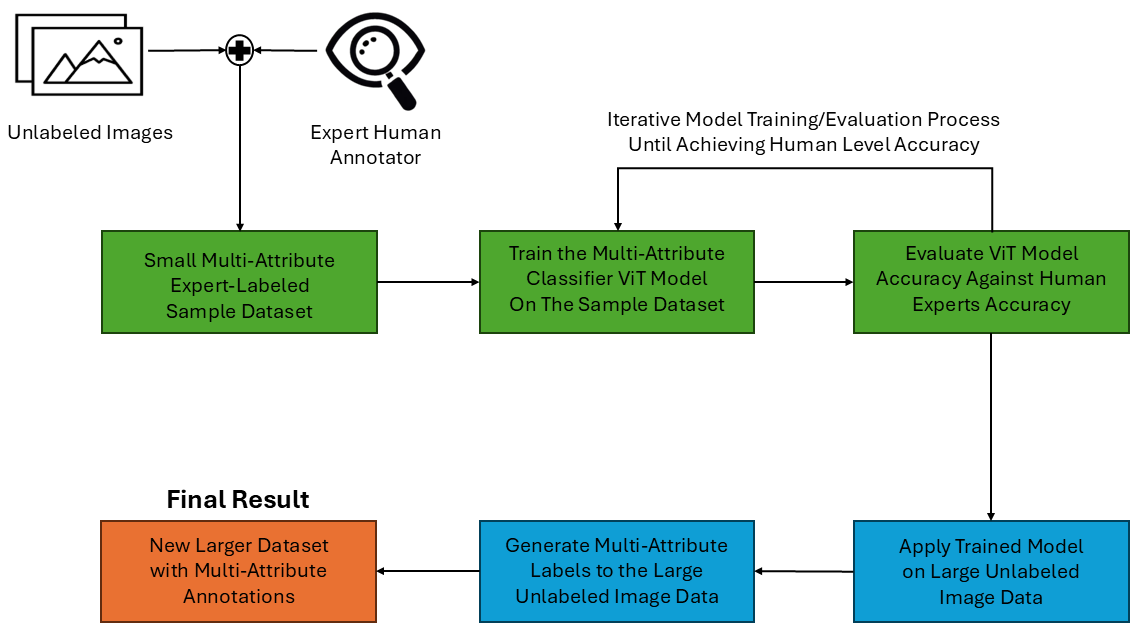}
    \caption{AttriGen Framework Architecture}
    \label{fig:AttriGen}
\end{figure}
As illustrated in Figure~\ref{fig:AttriGen}, AttriGen follows a bootstrapping approach that begins with a small expert-labeled dataset containing multi-attribute annotations. This seed dataset trains a ViT model that undergoes rigorous evaluation against human expert performance. Through iterative refinement, the model achieves near-human-level accuracy, at which point it becomes qualified to generate trusted annotations for previously unlabeled images. This process creates a substantially expanded dataset with comprehensive multi-attribute labels at a fraction of the time and cost of manual annotation.

\subsubsection{AttriGen: Impact on Computer Vision} 
AttriGen advances cell microscopy analysis in several key dimensions. First, it democratizes access to richly annotated datasets, enabling researchers without extensive annotation resources to develop and validate attribute-based recognition systems. Second, it facilitates the exploration of complex relationships between morphological attributes and disease states that would otherwise remain unexamined due to data limitations. Finally, while demonstrated in hematological image analysis, the AttriGen approach establishes a generalizable paradigm for expanding attribute-based image classification datasets across diverse medical imaging domains, potentially accelerating the development of more interpretable and clinically relevant computer vision systems throughout healthcare.

\section{Conclusion}
\label{sec:conclusion}

This paper introduces AttriGen, a novel framework for automated multi-attribute annotation in computer vision, with particular focus on blood cell microscopy. By creating a dual-model architecture that combines a CNN and ViT, our approach achieves 94.62\% GAA on multi-attribute classification while establishing a new paradigm for efficient dataset annotation. The integration of two complementary datasets; PBC for cell type classification and WBCAtt for morphological attributes, enables comprehensive 12-attribute characterization (11 attributes + cell type) aligned with real-world clinical assessment protocols.

Our automated approach demonstrates remarkable time efficiency, reducing annotation time for 6,784 images to approximately 2.26 minutes compared to extensive manual effort that would require weeks. The performance gap of 1.48\% between AttriGen (94.62\%) and human experts (96.1\%) is a reasonable trade-off given the enormous time savings necessary in scenarios where rapid analysis of large datasets is prioritized. Additionally, our framework used high-quality seed datasets (WBCAtt and PBC) for initial training, and validation with pathological samples achieving clinical robustness. Future research should explore extension to pathological cells with abnormal morphologies, which are critical for leukemia and other hematologic disease diagnosis. Additionally, active learning techniques could further optimize the annotation process, and enhanced explainability methods could highlight influential image regions for specific predictions. \textbf{}

\bibliographystyle{IEEEtran}
\bibliography{main}

\end{document}